\documentclass{llncs}
\usepackage{makeidx}  
\usepackage{tikz}
\usepackage{formatfig}
\usetikzlibrary{shapes,arrows}
\usepackage{amsmath,bm,times}
\usepackage{amssymb}
\usepackage{graphicx}
\usepackage{wrapfig, booktabs}
\graphicspath{{pics/}{figs/}}

\usepackage{titlesec} 
\titleformat{\subsection}[runin]{\normalfont\bfseries}{\thesubsection}{3pt}{}

\newcommand{\shapeworks}{\textit{ShapeWorks} }

\newcommand{\reffig}[1] {Figure~\ref{fig:#1}}

\newcommand{\refsec}[1] {Section~\ref{sec:#1}}

\begin{document}
\frontmatter          
\pagestyle{empty}  
\mainmatter              
\title{DeepSSM: A Deep Learning Framework for Statistical Shape Modeling from Raw Images}
%
%
\author{Riddhish Bhalodia\inst{1,2} \and Shireen Y. Elhabian\inst{1,2,3} \and Ladislav Kavan\inst{2} \and \\ Ross T. Whitaker\inst{1,2,3} }
%
%
%
\institute{Scientific Computing and Imaging Institute, University of Utah
  \and
 School of Computing, University of Utah
 \and 
 Comprehensive Arrhythmia Research and Management Center,  University of Utah
}


\maketitle              

\begin{abstract}
\vspace{-18pt}
Statistical shape modeling is an important tool to characterize variation in anatomical morphology. Typical shapes of interest are measured using 3D imaging and a subsequent pipeline of registration, segmentation, and some extraction of shape features or projections onto some lower-dimensional shape space, which facilitates subsequent statistical analysis. Many methods for constructing compact shape representations have been proposed, but are often impractical due to the sequence of image preprocessing operations, which involve significant parameter tuning, manual delineation, and/or quality control by the users. We propose DeepSSM: a deep learning approach to extract a low-dimensional shape representation directly from 3D images, requiring virtually no parameter tuning or user assistance. DeepSSM uses a convolutional neural network (CNN) that simultaneously localizes the biological structure of interest, establishes correspondences, and projects these points onto a low-dimensional shape representation in the form of PCA loadings within a point distribution model. To overcome the challenge of the limited availability of training images with dense correspondences, we present a novel data augmentation procedure that uses existing correspondences on a relatively small set of processed images with shape statistics to create plausible training samples with known shape parameters. In this way, we leverage the limited CT/MRI scans (40-50) into thousands of images needed to train a deep neural net. After the training, the CNN automatically produces accurate low-dimensional shape representations for unseen images. We validate DeepSSM for three different applications pertaining to modeling pediatric cranial CT for characterization of metopic craniosynostosis, femur CT scans identifying morphologic deformities of the hip due to femoroacetabular impingement, and left atrium MRI scans for atrial fibrillation recurrence prediction.



\end{abstract}


\section{Introduction}
\label{sec:intro}
\vspace{-8pt}

Since the pioneering work of D'Arcy Thompson \cite{thompson1942growth}, statistical shape models (SSM), also called morphological analysis, have evolved into an important tool in medical and biological sciences. A classical approach to comprehend a large collection of 2D or 3D shapes is via landmark points, often corresponding to distinct anatomical features. 
More recently, shape analysis for medical images is conducted with dense sets of correspondences that are placed automatically to capture the statistics of a population of shapes \cite{styner2006spharm,cates2007shape,davies2002MDL} or by quantifying the differences in coordinate transformations that align populations of 3D images \cite{beg2005computing}.  
The applications of these statistical shape methods are diverse, finding uses in orthopedics \cite{harris2013cam}, neuroscience \cite{greig2001brain}, and cardiology \cite{cates2013afib}.  

The goals of these kinds of analyses vary. In some cases, the analysis may be toward a clinical diagnosis, a task that might lend itself to a detection or classification strategy, which may bypass the need for any explicit quantification of shape. However, in many cases the goals include more open-ended questions, such as the formulation or testing of hypotheses or understanding/communicating pathological morphologies. Furthermore, training a state-of-the-art classifier for a specific disease would typically require (on the order of) thousands of samples/images for training, which becomes a significant burden for many clinical or biological applications, especially those involving human subjects.

Therefore, in this paper we address the problem of generating a rich set of shape descriptors in the form of PCA loadings on a shape space and an associated set of dense (i.e., thousands) correspondence points. The goal is to design a system that bypasses the typical pipeline of segmenting and/or registering images/shapes and the associated optimization (and associated parameter tuning)---and instead produces shape information directly from images via a deep (convolutional) neural network. This shape information can then be used to study pathologies, perform diagnoses, and/or visualize or study properties of populations or individuals.

Another contribution of this paper is the overall system architecture (and the demonstration of its efficacy on cranial, left atria and femur morphologies), which provides a blueprint for building other systems that could be built/trained to perform image-to-shape analyses with relative ease.  Another contribution is the particular strategy we have used for training, which relies on a conventional shape analysis on a relatively small set of images to produce a very large training/validation data set, sufficient to train a convolutional neural network (CNN).  
\vspace{-6pt}
\section{Related Work}
\vspace{-8pt}
%
The proposed system learns the projection of images onto a shape space,
which is built using correspondences between surfaces. Explicit
correspondences between surfaces have been done using geometric
parameterizations \cite{RTW:Sty2000,RTW:Dav2002} as well as functional
maps \cite{ovsjanikov2012functional}. In this work, we rely on a
discrete, dense set of samples, whose positions are optimized across a
population to reduce the statistical complexity of the resulting model. The resulting point sets can be then turned
into a low-dimensional shape representation by principal component
analysis (PCA), as in the method of point distribution models (PDMs)
\cite{RTW:Gre91}. For this optimization of correspondences, we use the open-source \textit{ShapeWorks} software \cite{cates2007shape,cates2017shapeworks}, which requires extensive pre-processing of input 3D images including: registration, segmentation, anti-aliasing (including a topology-preserving smoothing) and
conversion to a signed distance transform. These image processing
steps require well-tuned parameters, which, in practice, precludes a
fully automatic analysis of unseen scans.

Also related is the work on atlas building and computational anatomy
using methods of deformable registration (e.g., diffeomorphisms derived
from flows) \cite{beg2005computing}. Here, we pursue the
correspondence-based approach because many applications benefit from
explicit correspondences, exact matching of surfaces, and modes of
variation and shape differences that can be easily computed and
visualized for the surfaces under study (e.g., \cite{zachow2015computational}).  
While DeepSSM may be relevant for such registration-based
methods, such an approach would likely build on the many proposed
neural-net solutions to image registration
\cite{sokooti2017reg}.

DeepSSM builds on various works that have applied
convolution neural networks (CNNs) to 3D images \cite{lecun1998cnn}. More recently, deep learning is being generalized to mesh-based representations, with applications e.g., in shape retrieval \cite{xie2017deepshape}.
While much has been done in detection \cite{zheng2015detection},
classification \cite{li2014classification}, and segmentation (e.g., pixel
classification) \cite{badrinarayanan2015segnet2,ronneberger2015unet}, more directly relevant is the
work on regression from 3D images.  For
instance, in \cite{huang2017heartnet} they regress the orientation
and position of the heart from 2D ultrasound. Another recent work \cite{milletari2017stats} demonstrates the efficacy of PCA loadings in regressing for landmark position, being used for ultrasound segmentation task. DeepSSM extends this idea to 3D volumes and an extensive evaluation using it is performed on different datasets. DeepSSM proposes a novel data-augmentation scheme to counter limited-data availability in medical imaging applications. Furthermore, we employ the use of existing shape modeling tools to generate point distribution model and leverage the shape statistics for direct prediction of general shape parameters using CNN.


\vspace{-8pt}
\section{Methods}
\label{sec:method}

\begin{wrapfigure}[14]{r}{3.4in} 
  \begin{center}
    \includegraphics[scale = 0.3, trim = 0mm 0mm 0mm 30mm]{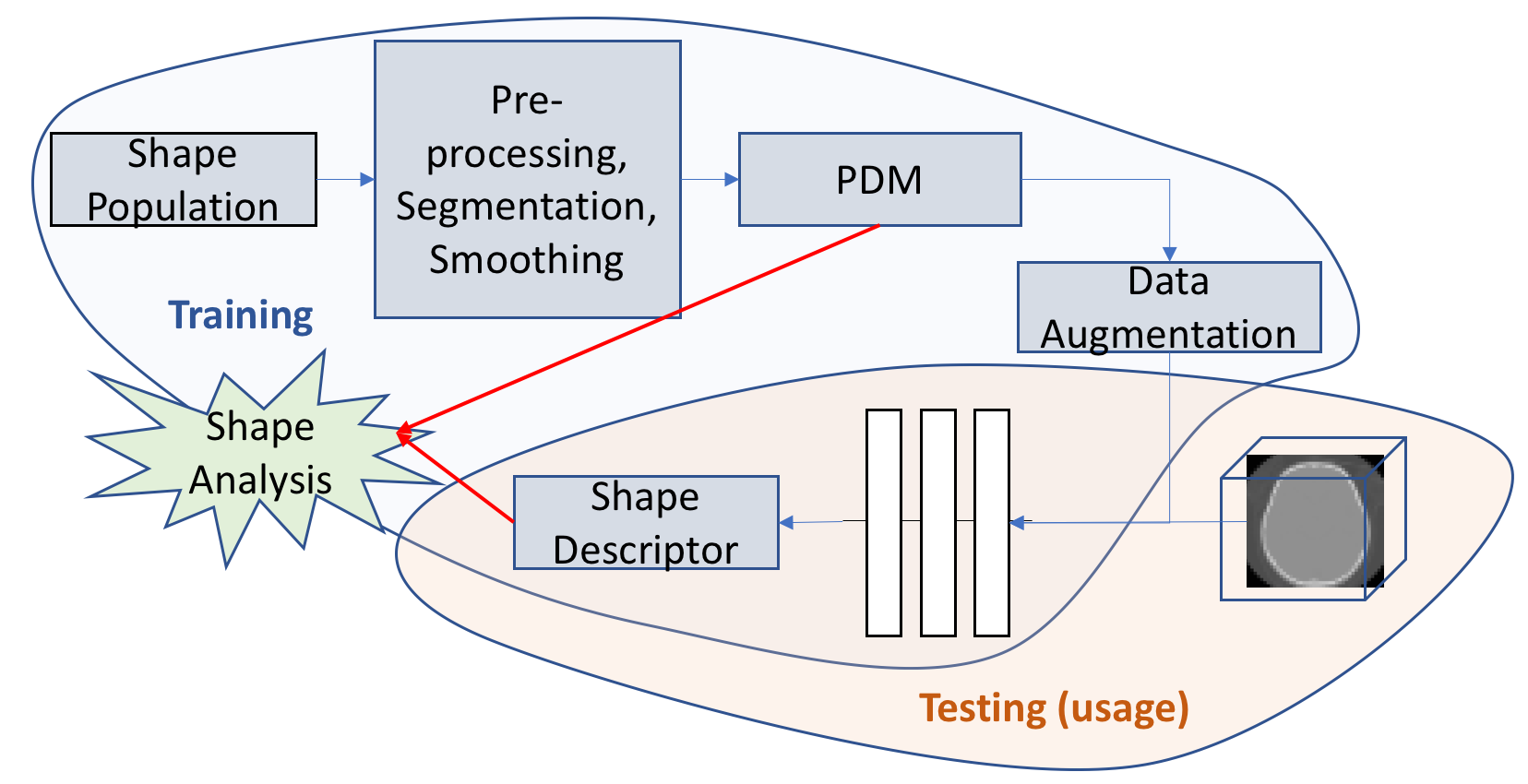}
    \caption{ Illustration of how DeepSSM is trained and used for getting shape descriptors for shape analysis directly from images.}
    \label{fig:workflow}
  \end{center}
\end{wrapfigure}

DeepSSM, unlike standard statistical shape modeling methods, is not a generative framework. It focuses on minimal pre-processing and direct computation of shape descriptors from raw images of anatomy that can be further used for shape analysis applications; some of which are described in the results section. \reffig{workflow} illustrates the training and usage of DeepSSM. In this section, we outline the data augmentation procedure, the CNN architecture and the learning protocols.

\setlength{\intextsep}{0pt}
\setlength{\columnsep}{5pt}
\setlength{\intextsep}{12pt plus 2.0pt minus 2.0pt}
\setlength{\columnsep}{10pt}

 
\subsection{Training Data Augmentation.}\label{sec:training}
\begin{figure}[htb]
     \centering
     \vspace{-10pt}
     \includegraphics[width=\textwidth]{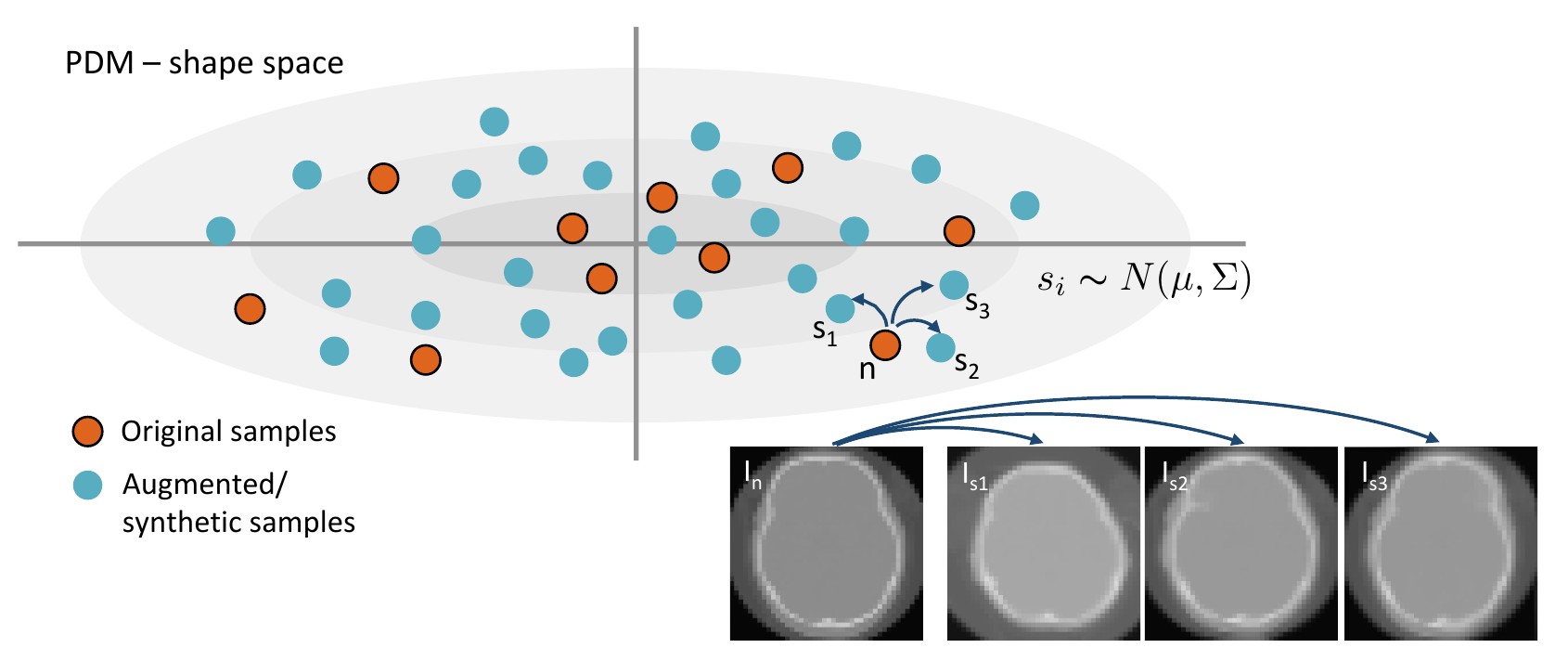}
     \vspace{-6pt}
     \caption{ Top: Shapes given by the original input CT scans (red dots) are augmented by sampling in the PDM shape space, from a normal distribution.  Bottom-right: The resulting correspondences are used to transform original images of nearby samples (with a TPS warp) to create new images with known shape parameters.}
     \vspace{-12pt}
     \label{fig:comparative-visual}
 \end{figure}

\vspace{-10pt} 
We start with a dataset with about 40-50 data samples, which are either CT or MRI images. These images are rigidly registered and downsampled to make the CNN training times manageable on current hardware, while still allowing for morphology characterization.
Because a set of 40-50 data samples is not large enough for CNN training, we propose a new data augmentation method. First, we construct a statistical shape model from surfaces extracted from the original data. We place a dense set of 3D correspondence points on each of the shapes using \shapeworks software \cite{cates2007shape,cates2017shapeworks}, even though any method of producing a PDM is applicable. We reduce this high dimensional PDM to $M$ dimensions using PCA, producing $M-$dimensional ``loading vectors'', where $M$ between 10-15 is usually sufficient to capture 99\% of the data variability. This corresponds to a multivariate ($M-$dimensional) Gaussian distribution, $\mathcal{N}(\mu, \Sigma)$. To generate a new synthetic image, we first draw a random sample $s \in \mathbb{R}^{M}$ from the $\mathcal{N}(\mu, \Sigma)$ distribution. This random sample $s$ corresponds to a statistically plausible shape.  To obtain a realistic 3D image associated with $s$, we find the closest example (denoted $n$) from input images. For this shape $n$, we already have a set of correspondences, $C_n$ and an associated image, $I_n$. We use the correspondences $C_n \leftrightarrow C_s$ to construct a thin-plate spline (TPS) warp \cite{bookstein1989principal} of $I_n$ to obtain a synthetic image $I_s$, which has the intensity profile of $I_n$ but the cranial shape $s$. The amount of TPS deformation is typically small, and using this method, 
we can generate thousands of new images that are consistent with the PCA space and intensity characteristics of the original dataset (see \reffig{comparative-visual}). We also employ an add-reject strategy to prevent extreme outliers from being created. In particular, we find the nearest neighbour of each generated sample from the original shapes and we reject the sample if the Euclidean distance between the two shapes exceeds a specified threshold.\\ \\

\subsection{Network Architecture.} 
\begin{figure}[htb]
    \centering
    \vspace{-8pt}
    \includegraphics[width=\textwidth]{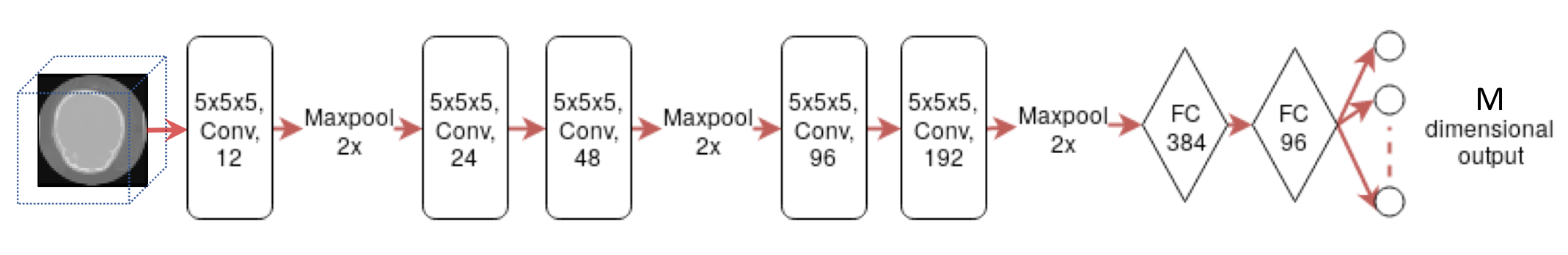}
    \vspace{-7pt}
    \caption{The architecture of the CNN network.}
    \vspace{-15pt}
    \label{fig:networkArch}
\end{figure}

We use a CNN architecture with five convolution layers followed by two fully connected layers to produce the output regression coefficients, see \reffig{networkArch}. The input to our network is a 3D image and the output is a set of ordered PCA loadings with respect to the shape space constructed in \refsec{training}. We found that in our setting, parametric ReLU \cite{he2015rectifiers} outperforms the traditional ReLU as a nonlinear activation function. We also perform batch normalization for all convolution layers. The weights of the network are initialized by  Xavier initialization \cite{xavier2010initialization}. 

\subsection{Learning Protocol.}\label{sec:learning}
We use 4000-5000 training data points and 1000-2000 validation images generated as described in \refsec{training}. We use TensorFlow \cite{abadi2016tensorflow} for constructing and training DeepSSM with a training batch size 10, which results in optimal saturation of the GPU (NVIDIA-Tesla K40c). The loss function is defined by taking $\mathbb{L}_2$ norm between the actual PCA loadings and the network output, and Adagrad \cite{duchi2011adagrad} is used for optimization. We use average root mean square error per epoch to evaluate convergence. We observed empirically that, in all datasets, this error becomes level after 50 epochs staying in range between $1.9-2.5$. Based on these observations we train our network for 60 epochs. 

\vspace{-8pt}
\section{Results and Discussion}


We apply DeepSSM on three different datasets pertaining to three different applications: (i) Pediatric cranial CT scans (ages : 5-15 months) for metopic craniosynostosis characterization, (ii) Left Atrium MRI data for prediction for the atrial fibrillation recurrence, and, (iii) Femur CT Data for the characterization of morphologic deformities of the hip due to femoroacetabular impingement. For each application, we divide the data into two categories, one which is used to generate the original PDM to be used for data augmentation, and the other data is completely quarantined and will be used to check the generalizablity of DeepSSM, we will refer to this data as ``unseen" data. We would like to stress that the unseen data is not part of the original data used for data augmentation or the PDM formation making it completely isolated. Further, we divide the data used for getting the PDM and it's accompanying augmented data into standard training, validation, and testing datasets. Another aspect to note is we perform a rigid ICP pre-alignment of all the images before computing it's PDM. It is important to note that DeepSSM is not an approach to discriminate between normal and pathological morphology, but an approach for reconstruction of shape representation from images that enables shape population statistics.

\begin{figure}[!h]
    \centering
    \vspace{-10pt}
    \includegraphics[width=\textwidth]{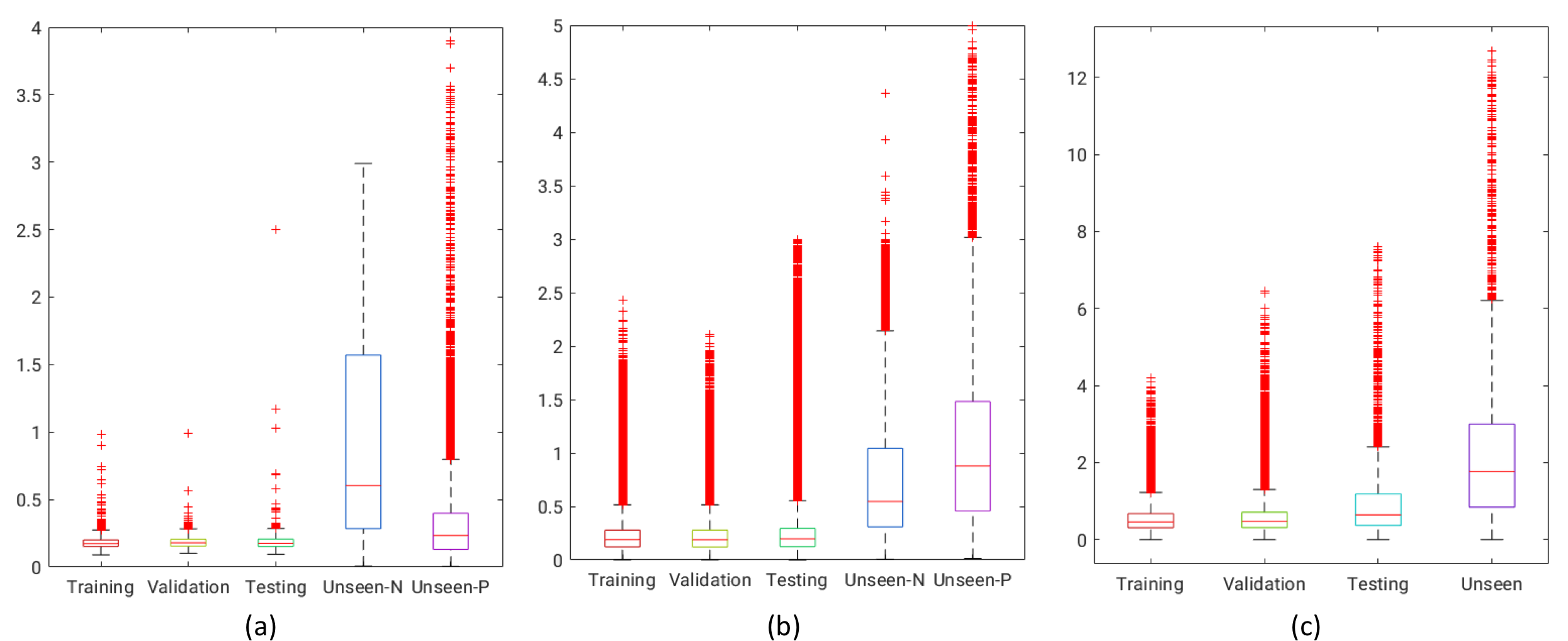}
    \caption{\textbf{Shape reconstruction errors in mm.} Each boxplot shows the error per-point per-shape in each category, for training, testing, and validation datasets. As ground truth correspondences are available, the error is simple Euclidean distance (in mm). For the unseen normal (unseen-N) and unseen pathological (unseen-P), the error is the minimum projection distance of the predicted point to original surface mesh (again in mm). (a) Metopic Craniosynostosis data, (b) Femur data, and (c) Left Atrium data.}
    \vspace{-10pt}
    \label{fig:allboxplots}
\end{figure}

\subsection{Metopic Craniosynostosis.}\label{sec:MC}
\begin{figure}[!h]
    \centering
    \vspace{-10pt}
    \includegraphics[width=\textwidth]{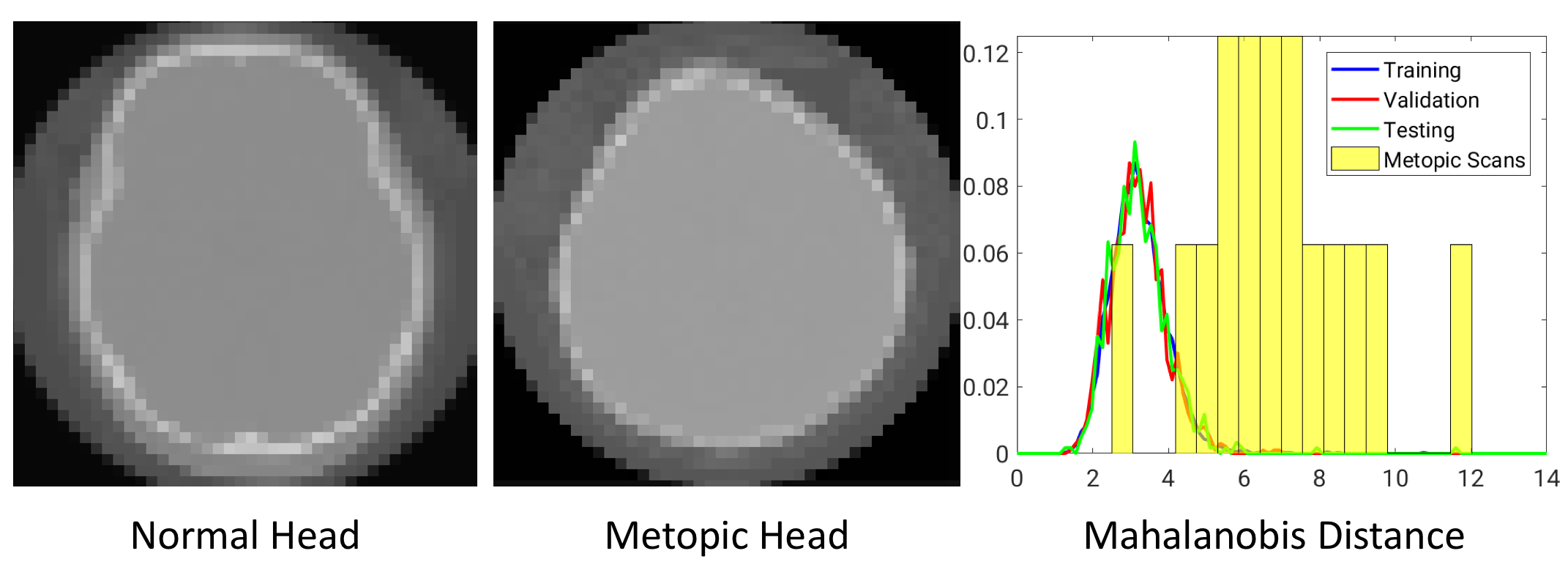}
    \caption{(Left) CT scan of a normal head. (Middle) CT scan of a head shape affected by metopic craniosynostosis. (Right) The histograms of Mahalanobis distance for training, validation and testing datasets and the metopic-heads dataset (yellow bars, no data augmentation was performed).}
    \vspace{-10pt}
    \label{fig:metopic}
\end{figure}

\vspace{-12pt} Metopic craniosynostosis is a morphological disorder of cranium in infants caused by premature fusion of the metopic suture, see \reffig{metopic}(metopic head). The characterization of the severity of the condition is hypothesized to be dependent on the deviation of skull shape from a normal phenotypical pediatric skull morphology. We aim to use DeepSSM to characterize this deviation. We use a dataset of 74 cranial CT scans of children with age between 5 to 15 months with 58 representing normal phenotypical skull morphology and 16 with metopic craniosynostosis, i.e., pathological skull deformities. 50 normative CT scans from the dataset were used in constructing our point distribution model, where each shape is represented by 1024 3D points; the vector of all these points is projected onto a 15-dimensional PCA subspace. We use this PDM to augment the data and train DeepSSM. We use the PCA loading predictions from a trained DeepSSM to reconstruct the 1024 correspondences and compare it with the original correspondences for obtaining the training, testing, and validation losses. To evaluate the accuracy of DeepSSM in predicting correspondences for unseen data, we use the remaining 8 CT scans of normal pediatric head shapes and 16 CT scans of children diagnosed with metopic craniosynostosis. We extract the outer skull surface from the unseen CT scans from a user-aided segmentation and render it as a triangle mesh using marching cubes. To account for an unknown coordinate system used in the unseen CT scan, we rigidly register this mesh to the 1024 3D correspondence points produced by DeepSSM from the raw CT scan. We then project these registered points to the surface of the mesh, these projection distances forms the error for evaluation on unseen data. The box plot representing the per-point per-shape Euclidean distance error, correspondence difference error for training, validation and testing data and the point to mesh projection error for the unseen cases, is shown in \reffig{allboxplots}(a). We observe that even though there is significant variability in skull shapes, the average error (across both data inclusive and exclusive to the data augmentation method) does not exceed 1mm. Our original CT scans were 1mm isotropic and they were downsampled by a factor of 4 making the voxel size to be 4mm, which means that DeepSSM predicts the correspondences with subvoxel accuracy.
It is encouraging that even though our initial shape space was constructed for only normal head shapes, DeepSSM generalizes well also to skulls with abnormal morphology resulting from metopic craniosynostosis. Next, we use this to explore an example application of automatic characterization of metopic craniosynostosis. We take 16 CT scans of pediatric patients diagnosed with metopic craniosynostosis and processed through DeepSSM, which produces their PCA loadings in the normative skull-shape space. We hypothesize that the skull shapes affected by metopic craniosynostosis will be statistically different from normal skull shapes. We compute the Mahalanobis distance between each of our head shapes (both normal and metopic ones) and our normative statistical shape model $\mathcal{N}(\mu, \Sigma)$. The histograms of these Mahalanobis distances for our datasets are shown in \reffig{metopic}. We can see that the histograms of training, validation, and testing images are closely overlapping, which is not surprising because these data sets correspond to normal phenotypical shape variations. However, the histogram of the metopic skull shapes indicates much larger Mahalanobis distance on average (yellow bars; the bars are wider and longer because no data augmentation was performed on the metopic craniosynostosis CT scans, and we have just 16 scans). The histograms of metopic-craniosynostosis and normal-skulls do overlap to some extent; this is indicative of mild cases of metopic craniosynostosis, which do not differ significantly from normal population and often do not require surgical intervention (unlike severe cases where surgery is often recommended \cite{mccarthy2012parameters}).

\subsection{Cam-type femoroacetabular impingement (cam-FAI).} 
\begin{figure}[!h]
    \centering
    \vspace{-8pt}
    \includegraphics[width=0.8\textwidth]{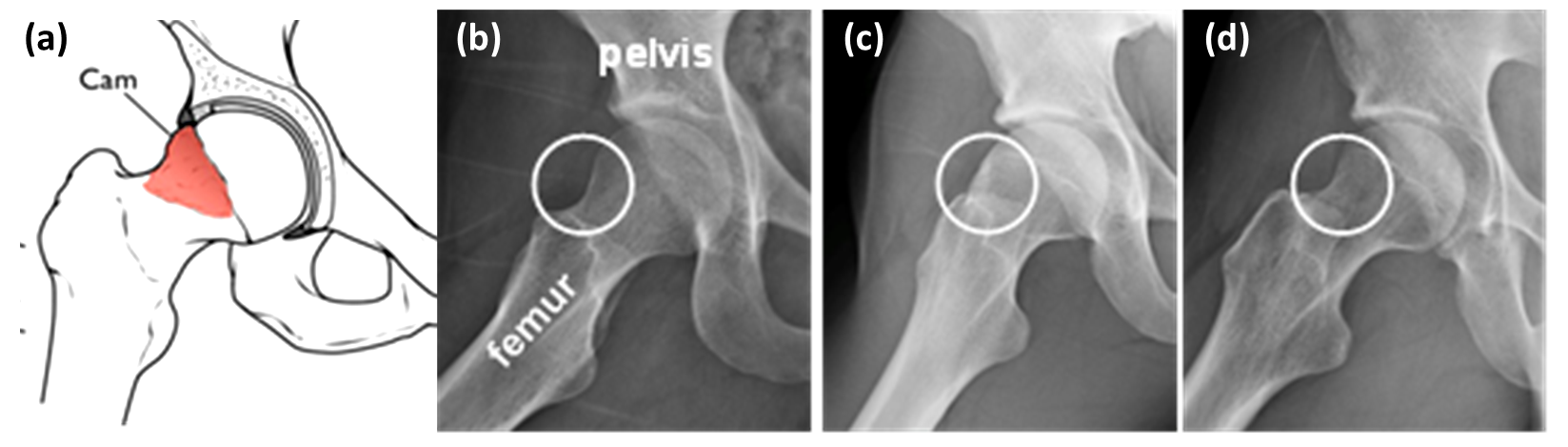}
    \caption{(a) Schematics of cam-FAI. Normal femur (b) compared to a cam femur (c); circles show location of deformity. (d) – cam FAI patient post-surgery.  Surgical treatment aims to remove bony deformities. }
    \vspace{-10pt}
    \label{fig:camFAI}
\end{figure}

\begin{figure}[!h]
    \centering
    \vspace{-8pt}
    \includegraphics[width=\textwidth]{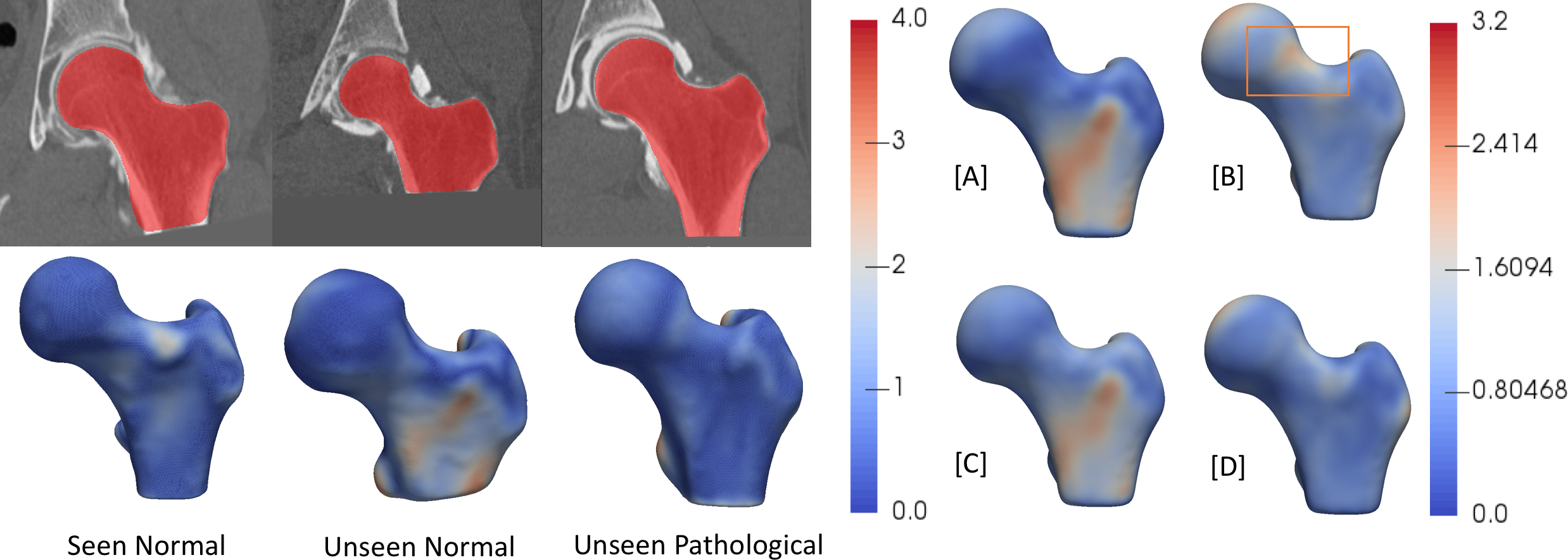}
    \caption{The left two rows represent the images (seen normal femur, unseen normal femur and unseen pathological femur) and the corresponding shape reconstruction error (Hausdorff distance in mm) interpolated as a heatmap on the original meshes. On the right : [A] mean error of unseen normal femurs overlayed on mean shape, [B] mean error of unseen pathological femurs overlayed on mean shape, [C] standard deviation of  error of unseen normal femurs overlayed on mean shape, [D] standard deciation of error of unseen pathological femurs overlayed on mean shape }
    \vspace{-10pt}
    \label{fig:FemurReconstruct}
\end{figure}

cam-FAI is a primary cause of hip osteoarthritis and is characterized by an abnormal bone growth of the femoral head (see \reffig{camFAI}). Statistical shape modeling could quantify anatomical variation in normal/FAI hips, thus providing an objective method to characterize cam-lesion and an anatomical map to guide surgical correction \cite{atkins2017quantitative}. We follow a very similar approach to that described in the Section \ref{sec:MC}.  Our dataset comprises of 67 CT scans with 57 femurs of normal patients and 10 pathological femurs. We start with 50 CT scans of femurs of normal patients which forms our control group. We reflect all the femurs to a consistent frame and then rigidly register them to a reference, the reflection is necessary as our data consists of both left and right femur bones. We use this set of 50 CT scans to form the PDM of 1024 3D correspondences, followed by its subsequent data augmentation. We again use 15 PCA loadings which captures $\sim 99\%$ of shape variability, and we train the DeepSSM to regress for these PCA loadings for 45 epochs. We use PCA loading predictions from the trained network to reconstruct the 1024 correspondences and compare it with the original correspondences. The box plot representing the Euclidean distance error (in mm) per-point per-shape is shown in \reffig{allboxplots}(b). For the unseen data (data which have no initial PDM on them), the error is again computed using the projection distance of the predicted correspondence from the original mesh. The femurs are also downsampled from 1mm isotropic voxel spacing by a factor of 4, making the voxel spacing 4mm. Our unseen data consists of the remaining 7 normal CT scans of femurs and 10 pathological femurs, We generate the correspondences from the PCA loading predicted by the DeepSSM and again evaluate the accuracy of the predictions using the maximal projected distance to the original mesh. The surface-to-surface distances for the unseen as well as seen scans are shown in \reffig{FemurReconstruct}. We want to evaluate the sensitivity of DeepSSM in predicting the subtle dysmorphology in the femoral head. As such, we compute the mean and standard deviation of the errors (i.e., surface to surface distances) for unseen normal and unseen pathological femurs and show them as a heatmap on a mean femur mesh. This is shown on the right in \reffig{FemurReconstruct}. A critical observation to note is that in \reffig{FemurReconstruct}[B], which is the mean error of the unseen pathological scans, the orange rectangle highlights the region of interest in characterization of the cam lesion. We observe that in this region, DeepSSM ---trained only on normal femurs--- results in a reconstruction error with sub-voxel accuracy,  which is not as accurate as some other (irrelevant to surgical treatment) regions of morphology. Aspect being stressed here is that pathological variation is not being captured by the training data, and hence the loss in reconstruction accuracy on pathologic cases. In particular, the network is learning a prior based on how a standard femur shape should look like and, it being trained on data augmented using a normative shape space, the pathological mode is not represented. Due to constraints in the data, we refrained from jointly modeling the initial PDM, which is essential if the pathological mode is to be captured using DeepSSM.

\subsection{Atrial Fibrillation.} 
\begin{figure}[!h]
    \centering
    \vspace{-10pt}
    \includegraphics[width=\textwidth]{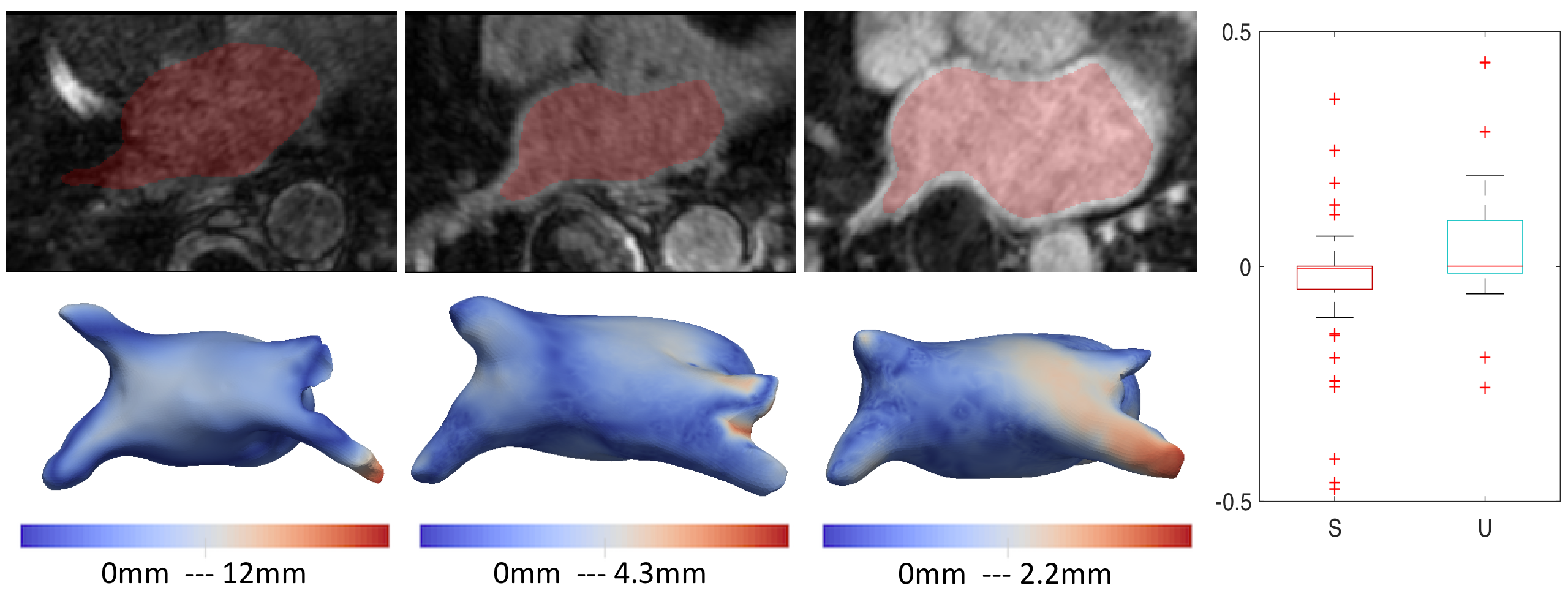}
    \caption{Bottom row represent representative sample of different image types in our database, going from worst to best (left to right). Top row represent shape derived from corresponding image using the proposed method, with a distance map overlay from particle modeling shape reconstruction. (Right) S: “Seen” Data
U: “Unseen“ Data : Boxplot for AF recurrence probability difference using PCA loadings using the PDM directly and those estimated by DeepSSM}
    \vspace{-10pt}
    \label{fig:LAreconst}
\end{figure}

Left atrium (LA) shape has been shown to be an independent predictor of recurrence after atrial fibrillation (AF) ablation \cite{cates2018afib}. Our dataset contains 100 MRIs of left atrium of paitents with AF. We start from 75 MRI scans from teh original dataset, and use them to form the initial PDM with 1024 points. We use 10 PCA components to capture 90\% of shape variability and use our data augmentation and train DeepSSM for 50 epochs. We use PCA loading predictions from the trained network to reconstruct the 1024 correspondences and compare it with the original correspondences. The box plot representing the Euclidean distance error (in mm) per-point per-shape is shown in \reffig{allboxplots}(c). The remaining 25 forms the unseen data (data which have no initial PDM on them), the error is again computed using the projection distance of the predicted correspondence from the original mesh. The MRI are downsampled from 1mm isotropic voxel spacing with a factor of 2 which leads to 2mm voxel spacing. We can see that DeepSSM performs poorly as compared to the other two applications on the unseen data. The reason for this is the huge variability in the intensity profile of the left atrium MRI, this is shown in \reffig{LAreconst}. We linearly scale the intensity range between 0-255 for training DeepSSM, but other then this there is no intensity equalization/correction is performed. It's encouraging that DeepSSM can still achieve an on average sub-voxel accuracy, and with smart subset selection, we believe the accuracy will increase substantially. Also in this analysis, we only use 10 PCA modes because, empirical observation of the other modes shows that they correspond to variations in the pulmonary veins, which is not important in AF recurrence prediction \cite{bieging2018left}, this also translates in the most error being concentrated in the pulmonary veins region \reffig{LAreconst}(leftmost). Furthermore, we want to see that how does DeepSSM work in predicting AF recurrence. We use the PCA loadings from the original PDM on the data and use them to perform multi-layer perceptron (MLP) regression against the AF recurrence data. We use this trained MLP and perform the same prediction, but now using the input data as the PCA loadings predicted using DeepSSM. We observe that the predicted recurrence probability using the PCA loadings from PDM and from DeepSSM are statistically same by T-Test with a confidence of 79.6\%. The recurrence probability difference from both inputs can be seen in \reffig{LAreconst}. We also perform a two one-sided test (TOST) \cite{schuirmann1987tost} for equivalence, we find that the recurrence prediction by DeepSSM and PDM PCA loadings are equivalent with a confidence of $88\%$ with the mean difference bounds of $\pm 0.1$.

\vspace{-8pt}
\section{Conclusion}
\label{sec:conclusion}
\vspace{-8pt}
DeepSSM provides a complete framework to extract low-dimensional shape representation directly from a shape population represented by 3D images. It provides a novel method to augment data from a small subset of images, and it's subsequent training. In contrast to previous methods which achieve similar functionality via a sequence of image processing operations and involve significant parameter tuning and/or user assistance, DeepSSM directly consumes raw images and produces a sub-voxel accurate shape model, with virtually no user intervention required for pre-processing the images unlike other shape modeling methods. We believe this functionality may enable new clinical applications in the future. We evaluate DeepSSM on both MRI as well as CT modalities, being applied to three different applications indicating that the framework is applicable to any collection of shapes.  Our preliminary analysis showing the efficacy of DeepSSM in pathology characterization for femoral heads and metopic craniosynostosis, even if it's trained on normal morphology, and opens up pathways to more detailed clinical analysis with DeepSSM on joint shape models. We hope that automatic shape assessment methods will contribute to new computerized clinical tools and objective metrics, ultimately translating to improved standards of medical care accessible to everyone.

\\

\noindent\textbf{Acknowledgment:} This work was supported by the National Institutes of Health [grant numbers R01-HL135568-01, P41-GM103545-19 and R01-EB016701]. This material is also based upon work supported by the National Science Foundation under Grant Numbers IIS-1617172 and IIS-1622360. Any opinions, findings, and conclusions or recommendations expressed in this material are those of the author(s) and do not necessarily reflect the views of the National Science Foundation.
The authors would like to thank the Comprehensive Arrhythmia Research and Management (CARMA) Center (Nassir Marrouche, MD), Pittsburgh Children's Hospital (Jesse Goldstein, MD) and the Orthopaedic Research Laboratory (Andrew Anderson, PhD) at the University of Utah for providing the left atrium MRI scans, pediatric CT scans, and femur CT scans, and their corresponding segmentations.

\vspace{-8pt}
\bibliographystyle{splncs03}
\vspace{-5pt}
\bibliography{cranio,rtw}

\end{document}